\documentclass{article}
\usepackage{ijcai21}

\usepackage{times}
\usepackage{soul}
\usepackage{url}
\usepackage[hidelinks]{hyperref}
\usepackage[utf8]{inputenc}
\usepackage[small]{caption}
\usepackage{graphicx}
\usepackage{amsmath}
\usepackage{amsthm}
\usepackage{multirow}
\usepackage{bm}
\usepackage{booktabs}
\usepackage{algorithm}
\usepackage{algorithmic}
\title{One Model for All Quantization:  \\  A Quantized Network Supporting Hot-Swap Bit-Width Adjustment}
\author{
Qigong Sun, Xiufang Li, Yan Ren, Zhongjian Huang, Xu Liu, Licheng Jiao, Fang Liu  \\
{\small Key Laboratory of Intelligent Perception and Image Understanding of Ministry of Education, International Research}\\
{\small Center for Intelligent Perception and Computation, Joint International Research Laboratory of Intelligent Perception}\\
{\small and Computation, School of Artificial Intelligence, Xidian University, Xi'an, Shaanxi Province 710071, China} \\
{\tt\small  xd\_qigongsun@163.com, xfl\_xidian@163.com, yanren@stu.xidian.edu.cn, huangzj@stu.xidian.edu.cn,  } \\
{\tt\small  xuliu361@163.com, lchjiao@mail.xidian.edu.cn, f63liu@163.com}
}

\begin{document}

\maketitle

\begin{abstract}
  As an effective technique to achieve the implementation of deep neural networks in edge devices, model quantization has been successfully applied in many practical applications.
  No matter the methods of quantization aware training (QAT) or post-training quantization (PTQ), they all depend on the target bit-widths.
  When the precision of quantization is adjusted, it is necessary to fine-tune the quantized model or minimize the quantization noise,
  which brings inconvenience in practical applications.
  In this work, we propose a method to train a model for all quantization that supports diverse bit-widths (e.g., form 8-bit to 1-bit)
  to satisfy the online quantization bit-width adjustment.
  It is hot-swappable that can provide specific quantization strategies for different candidates through multiscale quantization.
  We use wavelet decomposition and reconstruction to increase the diversity of weights,
  thus significantly improving the performance of each quantization candidate, especially at ultra-low bit-widths (e.g., 3-bit, 2-bit, and 1-bit).
  Experimental results on ImageNet and COCO show that our method can achieve accuracy comparable performance to dedicated models trained at the same precision.
\end{abstract}

\section{Introduction}

Quantization is one of the important technologies to promote the application of deep neural networks (DNNs) on resource-constrained edge devices.
It can achieve low power consumption and low latency inference in the case of low resource requirements.
Especially when the weights and activation values are quantized to extreme low-bit (e.g., 2-bit, and 1-bit),
the calculations can be replaced by logical operations (e.g., XNOR and Bitcount).
Quantizers convert infinite continuous floating-point values into finite discrete fixed-point values.
Commonly used quantizers can be categorized into three modalities: uniform quantizers, logarithmic quantizers and adaptive quantizers.
The obtained models are sensitive to specific quantizers and bit-widths.
In other words, the adjustment of quantization bit-width, range, or step size will easily cause severe performance degradation.
For example, Fig.\ 1(a) shows the ImageNet classification accuracies of the 8-bit ResNet-18 model tested on other bit-widths.
Obviously, when the quantization bit is less than 3, the accuracy will be reduced to about 0\%.

\begin{figure}[t]
    \setlength{\abovecaptionskip}{0.2cm}
    \setlength{\belowcaptionskip}{-0.2cm}
	\centering
	\includegraphics[width=3.3in]{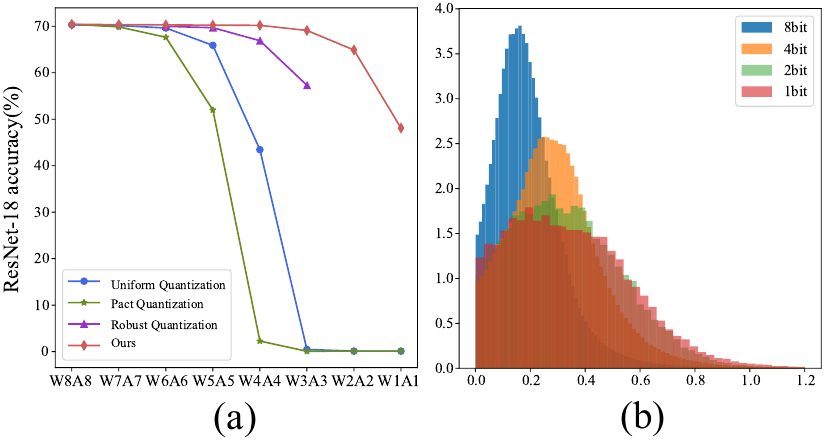}
	\caption{\small{
    (a) The ImageNet classification accuracies of quantized ResNet-18 models tested on multiple bit-widths.
    (b) Statistical distributions of activation values in ResNet-18 under different quantization bit-widths.
    }}
\end{figure}

More than 100 companies (e.g., Xilinx, Tesla, and Google) are currently working on DNN inference chips \cite{reddi2020mlperf},
and each chip has its own quantizer implementation.
In order to adapt to various quantization strategies, robust quantization, which can make the model suitable for more quantizers, has increasingly attracted the interest of researchers.
\cite{alizadeh2020gradient} and \cite{shkolnik2020robust} introduce some regularization terms (e.g., $l_1$-norm of gradients and Kurtosis regularization of weights)
to improve their quantization robustness.
Fig.\ 1(b) shows the distributions of activation values in ResNet-18 under different quantization bit-widths.
Obviously, the distribution of high-bits and low-bits are significantly different (e.g., 8-bit vs. 1-bit).
Therefore, when the span of the candidate space is large (e.g. 8-bit, 4-bit, 2-bit, and 1-bit),
the previous robust quantization strategies are highly unstable and lead to severe performance degradation.
For example, when quantizing ResNet-18 with robust quantization \cite{shkolnik2020robust} to 3-bit, the model leads to 13\% reduction in accuracy.
Robust quantization methods appear powerless under extreme quantization situations (e.g., 2-bit and 1-bit).
Therefore, it is particularly important to increase the diversity of weights and activation values to improve the performance under
low-bits quantization.

There are many practical applications where the quantization strategies (e.g., quantization bit-width, range, and step size)
need to be dynamically adjusted during the execution of the device.
For example, we can adjust the quantization bit-width according to the battery power dynamically, which is beneficial to preserve battery life by reducing power consumption.
However, no matter the methods of post-training quantization (PTQ) \cite{banner2019post} or quantization aware training (QAT) \cite{esser2019learned},
it is necessary to fine-tune the quantized model or minimize the quantization noise when the quantizer and bit-width are adjusted.
Cold-swap quantization bit-width adjustment process will take a significant amount of time, as shown in Fig.\ 2(a).
Therefore, there is an urgent need for a model that can support hot-swap bit-width adjustment in practical applications.


In order to obtain a quantized model that can support online adjustment of quantization bit-width, we propose a multiscale quantization method.
This method focuses on the multiscale quantization of activation values and weights.
By introducing a few hyper-parameters, we can obtain a model that provides all candidates (e.g., from 8-bit to 1-bit) with unique parameters through dynamic quantization training.
Our contributions can be concluded as follows:
\begin{itemize}
\setlength{\itemsep}{0pt}
    \item
    We present a hot-swappable mechanism that can adjust the bit-width during the operating period of the device without any optimization and fine-tuning,
    so as to meet the needs of online bit-width adjustment.
	\item
    Our model can support a wider range of bit-widths (e.g., from 8-bit to 1-bit) to support more practical applications,
    while maintaining the comparable performance to dedicated models trained at the same precision.
    \item
    By introducing the wavelet decomposition and reconstruction of the weight matrix, our model can generate specific weights for different quantization candidates.
    \item
    We conduct extensive experiments on benchmark datasets (ImageNet and COCO) to demonstrate the effectiveness of the proposed algorithm,
    which shows that our method are more effective than other counterparts under similar constraints.
\end{itemize}

\section{Related works}

\subsection{Model Quantization}
Model quantization usually quantizes full-precision parameters to low-bit to achieve model compression and acceleration.
The commonly used quantizers can be divided into uniform quantizers (e.g., PACT \cite{choi2018pact}, Dorefa-Net \cite{zhou2016dorefa}, QIL \cite{jung2019learning} and MBN \cite{sun2019multi}), logarithmic quantizers (e.g., LogQuant \cite{miyashita2016convolutional}, and INQ \cite{zhou2017incremental}) and adaptive quantizers (e.g., AdaBits \cite{jin2019adabits} and APoT \cite{li2019additive}).
They use different policies to realize the mapping from floating-point values to fixed-point values.
The mixed-precision quantization such as \cite{cai2020rethinking,uhlich2019mixed,sun2021effective} can match the sensitivity of each layer in DNNs
with  appropriate combination of quantization bit-widths, and it can achieve better results under the same constraints.
In practical applications, we can also choose the quantization method according to whether we need retraining.
PTQ methods, which are dedicated to reducing the noise of quantization,
are suitable for those scenarios where the training environment and dataset are not available at all \cite{banner2019post,nahshan2019loss}.
On the contrary,
QAT methods require dataset for training a more effective quantized model \cite{esser2019learned,choi2018pact}.

\subsection{Robust Quantization}

The quantized models are very sensitive to change in quantization strategies and adversarial noise.
In order to adapt to the practical applications, robust quantization has become an important research direction.
The common methods improve the robustness by restricting the networks or weights.
\cite{lin2019defensive} by controlling the Lipschitz constant of the network to alleviate the expansion of the adversarial noise in the quantized model.
\cite{alizadeh2020gradient} uses $l_1$-norm of gradients to control the maximum first-order induced loss thereby improving its robustness.
\cite{shkolnik2020robust} introduces a Kurtosis regularization term to uniformize the distribution of weights and improve the quantization robustness.
However, the above methods ignore the difference between the distribution of model weights and activation values under different bit-widths, which leads to performance degradation.
\cite{yu2019any}
makes the model be flexible in any numerical precision during inference by introducing multi-branch BatchNorm layer and dynamic model-wise quantization during training process.
However, it only considers the diversity of activation values but ignores the diversity of weights, which makes the accuracies of the model obviously decrease at high-bit quantization.

\subsection{Wavelet}

As a powerful time-frequency analysis tool, wavelet \cite{mallat1989theory,daubechies1992ten} can decompose the original signal/image into different frequency components.
This technique is often used for signal processing \cite{mallat1996wavelet,szu1992neural}.
Wavelet transform is also used as a tool for image content information analysis.
Discrete wavelet transform (DWT) can decompose the image into different frequency interpretations,
while inverse discrete wavelet transform (IDWT) can reconstruct multiple frequency components into the original image.
With the development of DNNs, wavelet transform has several attempts to combine the classical signal processing and deep learning methods,
such as image denoising \cite{kang2018deep,liu2020densely}, super resolution \cite{liu2018multi}, classification \cite{li2020wavelet,liu2020c}, segmentation \cite{li2020wavesnet}, facial aging \cite{liu2019attribute}, style transfer \cite{yoo2019photorealistic},  remote sensing image processing \cite{duan2017sar}, etc.
It is often used as the tool of data pre-processing, post-processing, feature extraction, and sampling operators in DNNs.
\cite{sun2021mwq} decomposes original data into multiscale frequency components by wavelet transform, and then quantizes the components of different scales, respectively.

\begin{figure*}[!h]
    \setlength{\abovecaptionskip}{0.1cm}
    \setlength{\belowcaptionskip}{-0.3cm}
	\centering
	\includegraphics[width=6.4in]{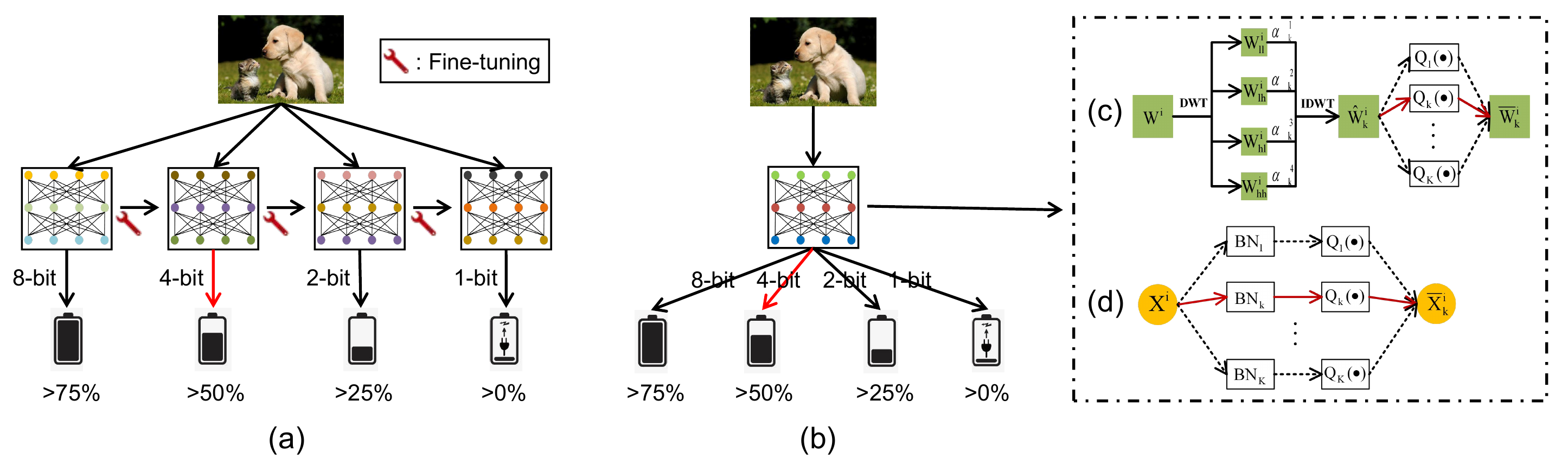}
	\caption{\small{
    (a) Cold-swap quantization bit-width adjustment process.
    When adjusting the bit-width of the quantized model, we need to pause the original process and fine-tuning it to get a new model.
    (b) Hot-swap quantization bit-width adjustment process.
    We can adjust the quantization bit-width by selecting different hyper-parameters without pausing the process.
    (c) Multiscale quantization of weights.
    (d) Multiscale quantization of activation values.
    }}
\end{figure*}

\section{Approach}

In order to meet the practical applications of hot-swap quantization bit-width adjustment,
we need to get a quantized model that can support multiple bit-widths at the same time.
Based on the multiscale analysis theory of wavelet decomposition and reconstruction,
we propose a multiscale quantization method to redefine a universal model, which can provide different quantization parameters for different candidates.
Assuming the weights of this model as $\mathbf{W}$ and the candidate space as $K$, we then can formalize the problem as
\setlength{\abovedisplayskip}{4pt}
\setlength{\belowdisplayskip}{4pt}
\begin{eqnarray}
\min_{\mathbf{W},\bm{\theta}}\sum _{k \in K} \mathcal{L}_{val}(\mathcal{T}(\mathbf{W},\bm{\theta}_k))
\end{eqnarray}
where $\mathcal{T}(\,)$ denotes the inference process, $\bm{\theta} _k$ denotes the hyper-parameters for $k$-bit quantization.
The overall training objective is to optimize $\mathbf{W}$ and $\bm{\theta}$ to make each candidate maintain the same level performance
as independently fine-tuning a model with the same bit-width.

By using this model, the quantization bit-width can be easily adjusted only by selecting different hyper-parameters, as shown in Fig.\ 2(b).
We use different wavelet reconstructed weights and hyper-parameters to increase the diversity of weights,
and use different Batch Normalization layers and quantization hyper-parameters to increase the diversity of activation values, as shown in Fig.\ 2(c) and (d).
For example, when the power of the edge device is less than 75\%, we can directly switch the quantization bit-width from 8-bit to 4-bit (the red line).
Note that the additional amount of hyper-parameters is negligible compared to the total number of network parameters.
Next, we will describe in detail the multiscale quantization of weights and activation values.



\subsection{Multiscale Quantization}
\subsubsection{Wavelet Decomposition and Reconstruction}

Wavelet is derived from multi-frequency analysis.
Low-pass filters and high-pass filters can decompose original data into various frequency components, which is called wavelet decomposition.
For example, we can decompose a tensor $\mathbf{X}$ into one low frequency component $\mathbf{X}_{ll}$ and three high frequency components $\mathbf{X}_{lh}, \mathbf{X}_{hl}$ and $\mathbf{X}_{hh}$ by single-level DWT.
The wavelet decomposition can be formulated as follows:
\setlength{\abovedisplayskip}{4pt}
\setlength{\belowdisplayskip}{4pt}
\begin{eqnarray}
\mathbf{X}_{c_0c_1}=(\downarrow2)(\mathbf{f}_{c_0c_1} \otimes  \mathbf{X})), \quad c_0,c_1 \in \{l,h\},
\end{eqnarray}
where $\otimes$ denotes the convolution operation, $(\downarrow2)$ denotes the downsample operation.
$\mathbf{f}_{ll}$ represents the low-pass filter and $\mathbf{f}_{lh},\mathbf{f}_{hl},\mathbf{f}_{hh}$ represent the high-pass filters,
which are determined by the wavelet base (e.g., \emph{Haar} and \emph{Daubechies}).
The decomposed components can be reconstructed to spatial domain through IDWT operation, as follows:
\setlength{\abovedisplayskip}{4pt}
\setlength{\belowdisplayskip}{4pt}
\begin{eqnarray}
\mathbf{X}=\sum_{c_0,c_1 \in \{l,h\}}\mathbf{f}_{c_0c_1}\oslash  (\uparrow2)\mathbf{X}_{c_0c_1},
\end{eqnarray}
where $\oslash$ denotes the deconvolution operation, $(\uparrow2)$ denotes the upsample operation.


\subsubsection{Multiscale Quantization of Weights}

The distribution and quantization strategies of weights are important factors that affect the performance of the quantized model.
Experiments show that a general model can not meet the quantization requirements of high-bits and low-bits at the same time, as shown in Fig.\ 1(a).
In order to adapt to various applications (e.g., 8-bit to 1-bit), we are committed to increasing the diversity of these factors.
Fig.\ 2(c) shows the schematic diagram of weights multiscale quantization.
We use the multiscale characteristics of wavelet decomposition and reconstruction to decompose the weights into multiple components,
and assign different scale factors to each component to increase the diversity of the weights after reconstruction.
We take the single-level wavelet transform as an example to express the main process:
\begin{eqnarray}
\mathbf{{W}}^{i}_{ll},\mathbf{{W}}^{i}_{lh},\mathbf{{W}}^{i}_{hl},\mathbf{{W}}^{i}_{hh} = \mathcal{DWT}(\mathbf{{W}}^{i}), \\
\mathbf{\hat{W}}^{i}_{k} = \mathcal{IDWT}(\alpha_{k}^{1}\mathbf{{W}}^{i}_{ll},\alpha_{k}^{2}\mathbf{{W}}^{i}_{lh},\alpha_{k}^{3}\mathbf{{W}}^{i}_{hl},\alpha_{k}^{4}\mathbf{{W}}^{i}_{hh}),
\end{eqnarray}
where $\mathcal{DWT}(\,)$ and $\mathcal{IDWT}(\,)$ represent the DWT and IDWT operations.
$\mathbf{\hat{W}}_{k}^{i}$ denotes the reconstructed weights of $i$-th layer for $k$-bit quantization.
$\mathbf{{W}}_{ll}^{i}$, $\mathbf{{W}}_{lh}^{i}$, $\mathbf{{W}}_{hl}^{i}$ and $\mathbf{{W}}_{hh}^{i}$ are the decomposed components,
$\alpha_{k}^{1}$, $\alpha_{k}^{2}$, $\alpha_{k}^{3}$ and $\alpha_{k}^{4}$ are learnable scale factors to each component.
$k\in K=\{1,2,3,4,5,6,7,8\}$ denotes the quantization bit-width, which can be dynamically changed in practical applications.

Based on the above multiscale transform, we can generate reconstructed weights $\mathbf{\hat{W}}_{k}^{i}$ with specific distribution
for different quantization candidates (e.g., form 8-bit to 1-bit).
Experience shows that quantization range and step size are different for different quantization precision.
In order to maintain the diversity of weights, we assign different quantization hyper-parameters to different reconstructed weights,
which can be formulated as follows:
\begin{eqnarray}
\mathbf{\bar{W}}^{i}_{k}=\mathcal{Q}_w(\mathbf{\hat{W}}^{i}_{k}, \mathbf{h}_k^w),
\end{eqnarray}
where $\mathbf{\bar{W}}_{k}^{i}$ denotes the quantized weights with $k$-bit quantization.
$\mathcal{Q}_w(\,)$ denotes the quantizer, in which the quantization hyper-parameters $\mathbf{h}_k^w$ are different for different bit-width candidates.

%
%
\subsubsection{Multiscale Quantization of Activation Values}

As feature maps extracted by neural networks, activation values play crucial roles in image recognition.
In order to achieve model compression and acceleration, the quantization of weights and activation values are needed.
The distributions of activation values under different bit-widths are obviously different, especially when bit-width is ultra-low (e.g., 2-bit and 1-bit),
as shown in Fig.\ 1(b).
In DNNs, Batch Normalization (BN) \cite{ioffe2015batch} is widely used to unify data and accelerate the convergence pace.
In order to maintain the diversity of activation values,
we match different BN layers and quantization hyper-parameters for different bit-width candidates,
which can be formulated as follows:
\begin{eqnarray}
\mathbf{\bar{X}}^{i}_k=\mathcal{Q}_a({ReLu}({BN}(\mathbf{X}^{i},\mathbf{B}_k)),\mathbf{h}_k^a)
\end{eqnarray}
where $\mathbf{\bar{X}}^{i}_k$ denotes the quantized $k$-bit activation values,
$\mathcal{Q}_a(\,)$ denotes the quantization function, in which $\mathbf{h}_k^a$ represents the quantization hyper-parameters.
$\mathbf{B}_k$ denotes the parameters of BN layer for $k$-bit quantization, respectively.




\subsection{Dynamic Quantization Training}

By introducing multiscale quantization, the quantized model can support multiple bit-widths to meet more practical applications.
Due to the obvious difference in the distribution of weights and activation values and hyper-parameters corresponding to different bit-widths, which makes joint network training difficult.
In order to solve this problem, we propose a dynamic network training method, which can dynamically adjust the quantization bit-width to stabilize the optimization process.
As a very sensitive factor in the quantizer, the initialization of the quantization hyper-parameters $\mathbf{h}$ and BN parameters $\mathbf{B}$ are particularly important in the networks.
Therefore, we apply a warmup procedure to provide suitable initialization of $\mathbf{B}$ and $\mathbf{h}$ for each candidate.
In order to improve the flexibility and robustness of the model, we dynamically change the quantization bit-width during the training process,
so that all candidates can participate in the network training.
The iterative procedure is shown in Algorithm 1.
Note that we can improve the performance of candidates by adjusting the sampling probability of specific bit-widths in the dynamic training process.

\begin{small}
\begin{algorithm}[!h]
\caption{Dynamic network training process.}
\label{alg:algorithm}
\textbf{Input}: The training dataset $\{(\mathbf{X}_i, \mathbf{y}_i)\}^N_{i=1}$;
Given candidate bit-widths $K=\{1,2,3,4,5,6,7,8\}$;\\
\textbf{Parameter}: Initialize weights $\mathbf{W}$ by a pre-trained model.
Initialize BN layer parameters $\mathbf{B}=\{\bm{\mu}_k,\bm{\sigma}_k,\bm{\gamma}_k,\bm{\beta}_k\}$,
scale factors $\bm{\alpha}=\{\bm{\alpha}_k^1,\bm{\alpha}_k^2,\bm{\alpha}_k^3,\bm{\alpha}_k^4\}$,
and quantization hyper-parameters $\mathbf{h}=\{\mathbf{h}_k^w,\mathbf{h}_k^a\}$. \\
\textbf{Output}: A quantized model supporting hot-swap bit-width adjustment.
\begin{algorithmic}[1] 
\STATE \textbf{Stage 1}: Warmup  $\mathbf{B}$ and $\mathbf{h}$.
\STATE Initialize $\bm{\alpha}$=\{1,1,1,1\}.
\FOR{$k$ in $K$}
\STATE Set quantization bit-width $k$.
\FOR{$i=1,\ldots,T$}
\STATE Calculate feed-forward loss.
\STATE Update $\mathbf{W}$, $\mathbf{B}$, and $\mathbf{h}$.
\ENDFOR
\ENDFOR
\STATE \textbf{Stage 2}: Dynamic training of quantized neural network.
\FOR{$epoch=1,\ldots,L$}
\FOR{$i=1,\ldots,T$}
\STATE Dynamically setting the quantization bit-width $k$.
\STATE Calculate feed-forward loss.
\STATE Update $\mathbf{W}$, $\mathbf{B}$, $\bm{\alpha}$, and $\mathbf{h}$.
\ENDFOR
\ENDFOR
\end{algorithmic}
\end{algorithm}
\end{small}

\begin{table*}[!t]
    \setlength{\abovecaptionskip}{0.2cm}
    \setlength{\belowcaptionskip}{-0.4cm}
    \footnotesize
	\centering
	\caption{Accuracy comparisons of ResNet-18 and ResNet-50 on ImageNet dataset.
    }
	\setlength{\tabcolsep}{1.2mm}{
		\renewcommand\arraystretch{1.1}
		\begin{tabular}{ccccccccccc}
			\hline
			\multirow{2}*{Models} & \multirow{2}*{Methods}    & \multirow{2}*{Size(M)} &\multicolumn{8}{c}{W/A-Bits}   \\ \cline{4-11}
            &&& 8/8   & 7/7    & 6/6    & 5/5    & 4/4    & 3/3    & 2/2    & 1/1  \\
			\hline
            \multirow{6}*{ResNet-18}
			&Fine-tuned Models             & 11.689    & 70.8 & 70.6 & 70.6 & 70.4 & 70.4 & 69.2 & 66.0 & 52.3     \\ \cline{2-11}
			&L1 Regularization \cite{alizadeh2020gradient}    & 11.689  & 69.92 & - & 66.39 & - & 0.22 & - & - & -        \\
            &L1 Regularization ($\lambda$ = 0.05) \cite{alizadeh2020gradient}  & 11.689  & 63.76 & - & 61.19 & - & 55.32 & - & - & -          \\
			&No Regularization \cite{shkolnik2020robust}    & 11.689  & - & - & 68.6 & 65.4 & 59.8 & 44.3 & - & -        \\
			&KURE Regularization \cite{shkolnik2020robust}  & 11.689  & - & - & 70.0 & 69.7 & 66.9 & 57.3 & - & -          \\
            &Ours [Haar]                  &  11.757    & \textbf{70.4} & \textbf{70.4}  & \textbf{70.3} &  \textbf{70.2}  & \textbf{70.2} & \textbf{69.1} & \textbf{64.9} & \textbf{48.1}     \\
            \hline
            \multirow{5}*{ResNet-50}
			&Fine-tuned Models  & 25.557  & 76.7 & 76.7 & 76.6 & 76.5 & 76.2 & 75.2 & 73.0  & 56.7      \\ \cline{2-11}
			&No Regularization \cite{shkolnik2020robust}    & 25.557  & - & - & 74.8 & 72.9 & 70.0 & 38.4 & - & -        \\
			&KURE Regularization \cite{shkolnik2020robust}  & 25.557  & - & - & \textbf{76.2} & \textbf{75.8} & 74.3 & 66.5 & - & -          \\
			&Any-Precision DNNs \cite{yu2019any}            & 25.557  & 73.5 & - & - & - & 71.3 & - & 65.2 & \textbf{55.3}         \\
            &Ours [Haar]                      & 25.931  & \textbf{75.6} & \textbf{75.6}  & {75.5} & {75.3}   & \textbf{75.3} & \textbf{74.8} & \textbf{71.9} & 53.5         \\
			\hline
		\end{tabular}
	}
	\label{tab:Resnettable}
\end{table*}

\subsection{Hot-Swap Bit-Width Adjustment}

Note that when adjusting the bit-width, the model weights need to perform certain operations to obtain the quantized weights for the new candidate.
These operations mainly include model loading, DWT, IDWT and quantization, as shown in Fig.\ 2(c).
For example, the bit-width adjustment process of ResNet-50 can be completed in 1.2s on CPU.
This process is hot-swappable, which is not perceived by users in practical applications.

\section{Experiments}

In order to verify the robustness, diversity and effectiveness of our method,
we conduct experiments on ImageNet and COCO benchmarks.
In this section, we first describe the details of our experimental implementations.
Then, the experimental results of our method are presented to compare with other methods.
Finally, we analyze the wavelet decomposition of the weights and ablation studies of our method.

\subsection{Implementation Details}

In the implementation, we can use convolution and deconvolution to achieve wavelet decomposition (DWT) and reconstruction (IDWT).
The low-pass filters and high-pass filters can be defined as convolution kernels.
For example, the filters of the corresponding 2D Haar wavelet are
\setlength{\abovedisplayskip}{4pt}
\setlength{\belowdisplayskip}{4pt}
\begin{small}
\begin{eqnarray*}
\mathbf{f}_{ll}=\frac{1}{2}\begin{bmatrix}
1\; & \;\; 1\\
1\; & \;\; 1
\end{bmatrix},
\mathbf{f}_{lh}=\frac{1}{2}\begin{bmatrix}
1 & 1\\
-1 & -1
\end{bmatrix},\\
\mathbf{f}_{hl}=\frac{1}{2}\begin{bmatrix}
1 & -1\\
1 & -1
\end{bmatrix},
\mathbf{f}_{hh}=\frac{1}{2}
\begin{bmatrix}
1 & -1\\
-1 & 1
\end{bmatrix}.
\end{eqnarray*}
\end{small}
\!\!Therefore, we can use addition, subtraction and shift operation to implement wavelet transform, which is very efficient in calculation.
A weight matrix ($1024\times 512\times 3 \times 3$) can be decomposed into four components ($512\times 256\times 3 \times 3$).
Here, we use PACT \cite{choi2018pact} as the quantizer to quantize weights and activation values.
For ImageNet, the first convolutional layer and the last fully-connected layer are quantized to $8$-bit, and the batch size for all the networks is set to $1024$.
We use SGD as the optimizer, in which the initial learning rate is $1 \times 10^{-2}$ and the weight decay is set to $1 \times 10^{-4}$.
For COCO detection, we use ResNet-50 as the backbone.
Our network is ﬁne-tuned by SGD for 16 epochs with the initial learning rate $1 \times 10^{-2}$ and the batch size of 32 on 8 Nvidia Tesla V100 GPUs.
The learning rate is decayed by a factor of 10 at epoch 4 and 8, respectively.


\subsection{Experimental Results}

\subsubsection{Image Classification}

ImageNet (ILSVRC2012) is a common large-scale dataset for image classification task, which has been widely used in academic researches.
It contains more than 1.2M training images and 50K validation images.
Here, we implement experiments with ResNet-18 and ResNet-50 on this dataset to illustrate the effectiveness of our method,
and compare the performance of our method with those of state-of-the-art methods, such as \cite{alizadeh2020gradient,shkolnik2020robust,yu2019any}.
For each network architecture, we obtain a model that supports multiple quantization candidates.

Table 1 shows the Top-1 accuracy comparisons of ResNet-18 and ResNet-50 on ImageNet dataset.
The first line denotes the fine-tuned models for each bit-width, which are inconvenient to obtain these models in practical applications due to the need for multiple fine-tuning.
\cite{alizadeh2020gradient,shkolnik2020robust} introduce regularization term to restrict the distribution of weights, thereby improving the robustness of quantized networks.
Obviously, when the quantization bit-width is low, the accuracies are greatly reduced, especially in the cases of extreme quantization (e.g., 2-bit, and 1-bit).
For example, the accuracy drops to 0.22\% at 4-bit quantization in `L1 Regularization', therefore,
this method is invalid when the quantization bit-width is less than 4.
Similarly, the effectiveness of `KURE Regularization' in the low bit-width cases are also insufficient.
Any-Precision DNNs \cite{yu2019any} focuses on the optimization of low-bits (e.g., 2-bit, and 1-bit) quantization,
while the accuracies of high-bits (e.g., 8-bit, and 4-bit) quantization decrease significantly.
Obviously, our method can achieve accuracy comparable to dedicated models trained at the same precision,
and the model size is slightly increased.
However, in the lower bit-widths (e.g., 3-bit, 2-bit, and 1-bit) cases, the accuracies of our models are significantly improved.
In particular, our method achieves 69.1\% ImageNet Top-1 accuracy on ResNet-18 model at 3-bit, outperforming its counterpart \cite{shkolnik2020robust} by a large margin of 11.8\%,
and the accuracy of our ResNet-50 model at 2-bit is higher 6.7\% than \cite{yu2019any}.

According to the characteristics of wavelet decomposition and reconstruction,
we provide the diversity of weight distribution through the combination of frequency components of different scales.
Taking the weights of `layer 4.0.conv1' in ResNet-18 as an example,  Fig.\ 3 shows the statistical distributions of reconstructed weights for 8-bit and 1-bit quantization.
This obvious distribution makes it possible for us to achieve accuracy comparable results to the fine-tuned model on 1-bit quantization.
\vspace{-0.2cm}
\begin{figure}[!h]
    \setlength{\abovecaptionskip}{0.1cm}
    \setlength{\belowcaptionskip}{-0.4cm}
	\centering
	\includegraphics[width=3.0in]{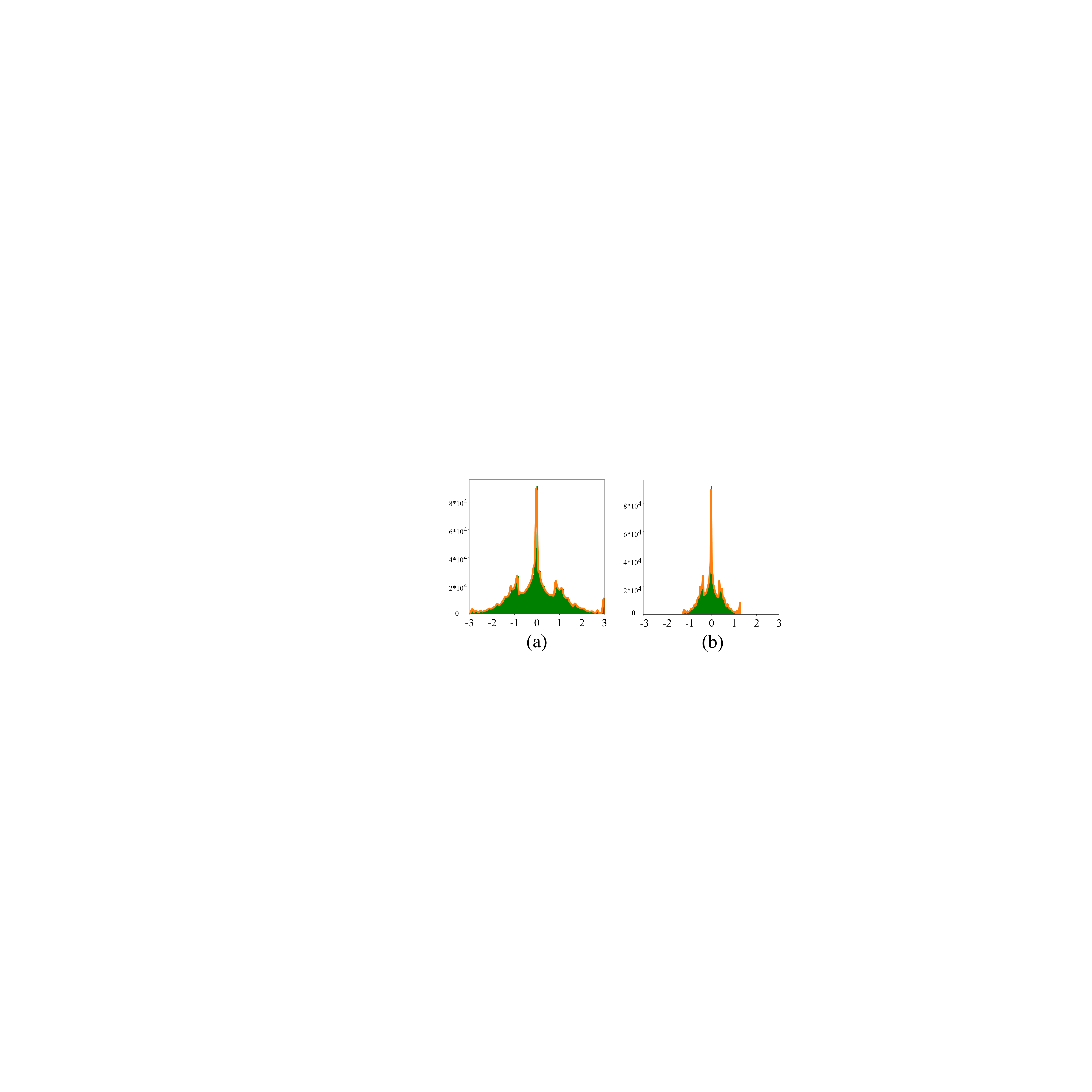}
	\caption{\small{
    (a) Statistical distribution of reconstructed weights for 8-bit quantization.
    (b) Statistical distribution of reconstructed weights for 1-bit quantization.
    }}
\end{figure}

\subsubsection{COCO Detection}

We further explore the effectiveness of our method for object detection and instance segmentation tasks on the COCO benchmark \cite{lin2014microsoft}.
This is one of the most popular large-scale benchmark datasets, which consists of images from 80 different categories and 115K images for training and 5K images for validation.
We use ResNet-50 as the backbone of Mask R-CNN \cite{he2017mask} to implement our quantized networks.
In training process, the quantization bit-widths of backbone are dynamic and other parameters remain at full-precision.
Four experiments are implemented in this section, where `Fine-tuned Models' denotes the results are fine-tuned for each bit-width.
For our method, we use \emph{Haar} as the wavelet base to decompose weights, and only one model can be obtained for all candidates.
The main difference between our experiments is the candidate space, e.g., the candidate space of `Ours$^2$' is \{3,4,5,6,7,8\}).

\begin{table}[!h]
    \setlength{\abovecaptionskip}{0.1cm}
    \setlength{\belowcaptionskip}{-0.3cm}
    \footnotesize
    \newcommand{\tabincell}[2]{\begin{tabular}{@{}#1@{}}#2\end{tabular}}
	\centering
	\caption{Results of Mask R-CNN on the COCO validation set.}
	\setlength{\tabcolsep}{1.1mm}{
		\renewcommand\arraystretch{1.1}
		\begin{tabular}{c|cc|cc|cc|cc}
			\hline
			\multirow{1}*{Methods} & \multicolumn{2}{c|}{\tabincell{c}{Fine-tuned \\ Models}} & \multicolumn{2}{c|}{Ours$^1$[Haar]} & \multicolumn{2}{c|}{Ours$^2$[Haar]}  & \multicolumn{2}{c}{Ours$^3$[Haar]} \\ \cline{1-9}
            Size(M)  & \multicolumn{2}{c|}{44.396} & \multicolumn{2}{c|}{44.715} & \multicolumn{2}{c|}{44.662}  & \multicolumn{2}{c}{44.609} \\ \cline{1-9}
            W/A-Bits & \textrm{d-AP} & $\textrm{s-AP}$  & $\textrm{d-AP}$ & $\textrm{s-AP}$ & $\textrm{d-AP}$ & $\textrm{s-AP}$ & $\textrm{d-AP}$ & $\textrm{s-AP}$ \\
            \hline
			 8/8     & 40.6   & 36.7  & 38.9   & 35.3   & 40.2   & 36.3  & 40.5   & 36.6 \\
			 7/7     & 40.1   & 36.4  & 38.8   & 35.2   & 40.1   & 36.2  & 40.5   & 36.6   \\
			 6/6     & 40.4   & 36.7  & 38.8   & 35.1   & 40.0   & 36.2  & 40.4   & 36.5  \\
             5/5     & 40.3   & 36.5  & 38.7   & 35.0   & 39.8   & 36.0  & 40.0   & 36.2  \\
			 4/4     & 39.9   & 36.0  & 38.0   & 34.4   & 39.2   & 35.6  & 39.5   & 35.9    \\
			 3/3     & 38.5   & 34.8  & 35.8   & 32.6   & 34.5   & 31.9  & -   & - \\
			 2/2     & 17.9   & 16.8  & 10.6   & 10.3   & -   & -  & -   & -  \\
			\hline
		\end{tabular}
	}
	\label{tab:Resnettable}
\end{table}

We report the standard COCO metrics for object detection and instance segmentation tasks, which are labeled as $\textrm{d-AP}$ and $\textrm{s-AP}$.
Experimental results are shown in Table 2.
Compared with the fine-tuned model, our model still has comparability in each candidate bit width, especially when the low-bits (e.g., 3-bit, and 2-bit) are not considered.
The effect of low-bit candidates on this task is obvious.
Taking into account 2-bit quantization as a candidate lead to high-bit performance reduction.

\subsection{Multiscale Analysis of Weights}

The weight matrix decomposition of DNNs has achieved great performance in model compression and acceleration \cite{wang2018packing}.
Weight matrix can be approximated as a weighted linear combination of basis filters.
In this paper, we use wavelet to decompose the weight matrix, and the four components represent the mapping of the weight matrix in different frequency domains.
The four components can reconstruct weights to participate in the calculation of networks.
Table 3 shows the classification accuracies of the weights reconstructed by different combinations on ImageNet,
in which $\mathbf{{W}}_{ll}$+$\mathbf{{W}}_{lh}$ represents the weights reconstructed only by $\mathbf{{W}}_{ll}$ and $\mathbf{{W}}_{lh}$.
From table 3, we can see that the low frequency component $\mathbf{{W}}_{ll}$ is still the most important component in the weights $\mathbf{{W}}$.
We can get 56.9\% and 66.3\% classification accuracies on ResNet-18 and ResNet-50 by using low frequency component $\mathbf{{W}}_{ll}$.
With the continuous addition of high frequency components, the accuracies are getting higher and higher.
Fig.\ 4 shows the statistical distributions of weights and decomposed components.
There are obvious differences in the distribution of low frequency component and high frequency components, especially the high frequency component $\mathbf{{W}}_{hh}$.
It is the diversity combination of these four components that makes weight diversity possible.

\begin{table}[!h]
    \setlength{\abovecaptionskip}{0.2cm}
    \setlength{\belowcaptionskip}{-0.2cm}
    \footnotesize
	\centering
	\caption{Accuracy comparisons of ResNet-18 and ResNet-50 with different weight components reconstruction on ImageNet.
    }
	\setlength{\tabcolsep}{2mm}{
		\renewcommand\arraystretch{1.2}
		\begin{tabular}{c|c|c|c}
			\hline
			Models  & $\mathbf{{W}}_{ll}$    & $\mathbf{{W}}_{ll}$+$\mathbf{{W}}_{lh}$   & $\mathbf{{W}}_{ll}$+$\mathbf{{W}}_{lh}$ +$\mathbf{{W}}_{hl}$  \\
            \hline
			ResNet-18 [Haar] &  56.9   &62.2   & 66.8             \\
			ResNet-50 [Haar] & 66.3   & 69.0    & 73.5             \\
			\hline
		\end{tabular}
	}
	\label{tab:Resnettable}
\end{table}


\begin{figure}[!h]
    \setlength{\abovecaptionskip}{0.1cm}
    \setlength{\belowcaptionskip}{-0.2cm}
	\centering
	\includegraphics[width=2.8in]{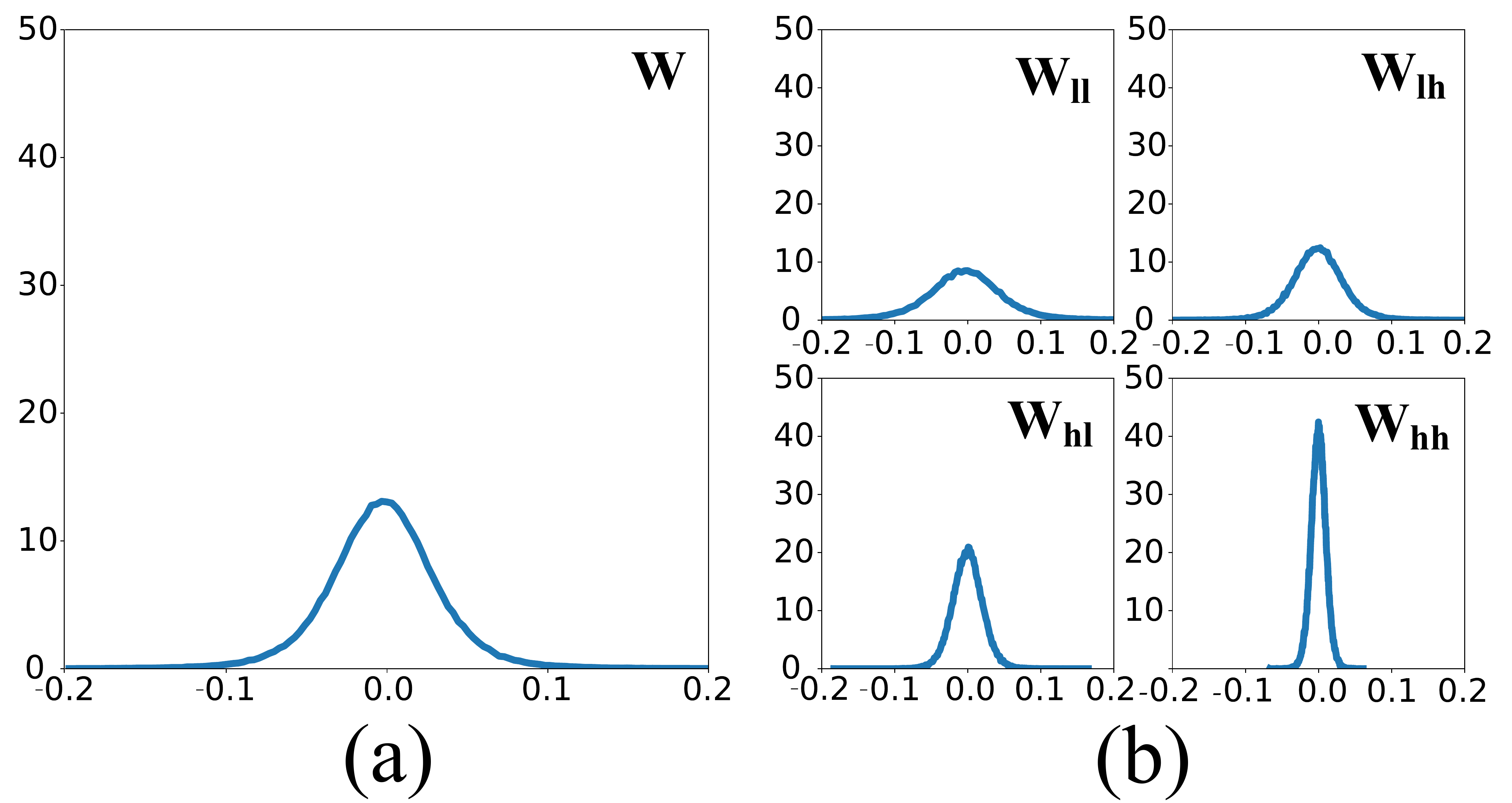}
	\caption{\small{
    (a) Statistical distribution of weights `layer3.0.conv2' in ResNet-18.
    (b) Statistical distributions of decomposed components.
    }}
\end{figure}

\subsection{Ablation Studies}

In this paper, we can obtain a model that supports multiple bit-widths quantization without any fine-tuning process.
In order to verify the effectiveness of those strategies,
we design the following ablation experiments on ResNet-18 to explore the impacts of using multiscale quantization of weights and activation values.

\begin{figure}[t]
    \setlength{\abovecaptionskip}{0.1cm}
    \setlength{\belowcaptionskip}{-0.2cm}
	\centering
	\includegraphics[width=2.6in]{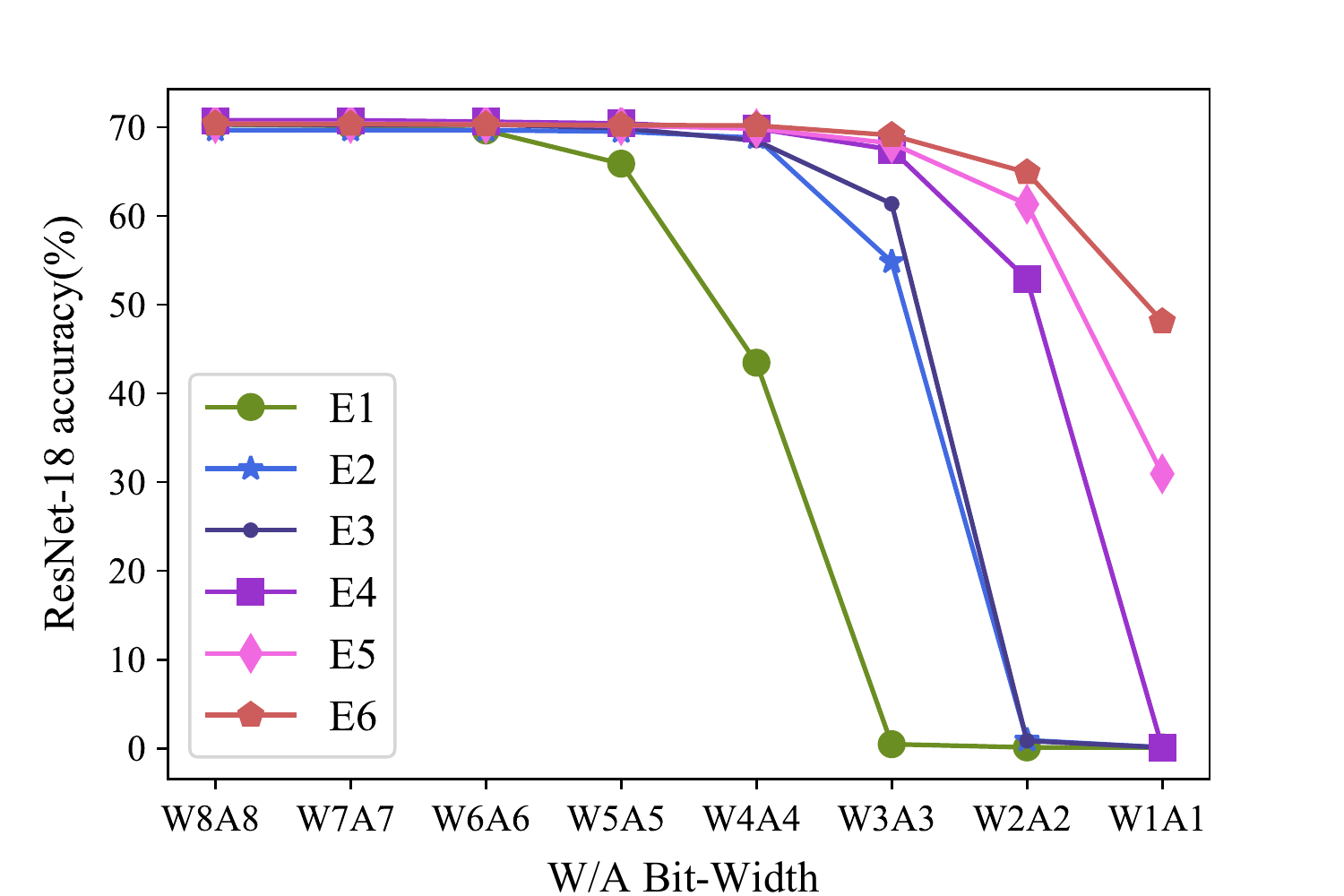}
	\caption{\small{The ablation experiments of ResNet-18 on ImageNet.
    }}
\end{figure}

\subsubsection{Effect of Dynamic Quantization Training}
Dynamic quantization training can dynamically adjust the quantization precision during the training process.
Thus, a variety of bit-widths can jointly constrain the optimization process.
Here, we use a pre-trained 8-bit quantization model to compare with a dynamic trained single scale model.
The experimental results are shown as \emph{E1} and \emph{E2} in Fig.\ 5.
It is obvious that dynamic quantization training can effectively improve the robustness of the model, especially for 4-bit and 3-bit quantization.

\subsubsection{Effect of Multiscale Quantization of Activation Values}

Multiscale quantization of activation values means that we provide different Batch Normalization layers and quantization  hyper-parameters (e.g., step size) for different quantization candidates.
In \emph{E3}, we provide different quantization scales with activation values for different candidates.
In \emph{E4}, we use multiple Batch Normalization layers on the basis of \emph{E3} to realize the specific normalization of different candidates.
When performing multi-bit quantization verification, the weights and their hyper-parameters are shared.
From Fig.\ 5, we can see that multiple Batch Normalization layers can improve the performance of 3-bit and 2-bit quantization, but it is still incapable for 1-bit quantization.


\subsubsection{Effect of Multiscale Quantization of Weights}
We are committed to analyzing the effect of multiscale quantization of weights.
On the basis of \emph{E4}, we add multiscale quantization and wavelet transform of weights as \emph{E5} and \emph{E6}.
As can be seen from Fig.\ 5, applying multiscale quantization for weights, different hyper-parameters are provided for different quantization candidates,
will lead to a significant effect for 1-bit quantization.
The existence of wavelet diversity enables our model to guarantee high-bit accuracies while allowing low-bit performance comparable to
dedicated models trained at the same precision.

\section{Concluding Remarks}

In this paper, we innovatively propose a multiscale quantization method,
which can train a quantized model supporting hot-swap bit-width adjustment to meet the needs of practical applications.
Our model can provide specific parameters for different quantization candidates,
and achieve comparable results in image classification, target detection and instance segmentation tasks.
Multiscale analysis provides more possibilities for weights diversity.
In the future, we will strive to achieve the rapid adjustment of quantization bit-width without training datasets.

{\small
\bibliographystyle{named}
\bibliography{ijcai21}
}

\end{document}